%% file: main.tex
\documentclass[10pt,twocolumn,letterpaper]{article}

\usepackage{cvpr}      % To produce the REVIEW version
\definecolor{cvprblue}{rgb}{0.21,0.49,0.74}
\usepackage[pagebackref,breaklinks,colorlinks,allcolors=cvprblue]{hyperref}

\def\paperID{*****} % *** Enter the Paper ID here
\def\confName{CVPR}
\def\confYear{2026}

\title{Global Multiple Extraction Network for Low-Resolution Facial Expression Recognition}

\author{Jingyi Shi\\jingyishicn@gmail.com}

\begin{document}
\maketitle
\input{sec/0_abstract}    
\input{sec/1_intro}

\input{sec/2_formatting}
\input{sec/3_finalcopy}
\input{sec/4}

\input{sec/5}

{
    \small
    \bibliographystyle{ieeenat_fullname}
    \bibliography{main}
}


\end{document}

%% file: sec/0_abstract.tex
\begin{abstract}
  Facial expression recognition, as a vital computer vision task, is garnering significant attention and undergoing extensive research. Although facial expression recognition algorithms demonstrate impressive performance on high-resolution images, their effectiveness tends to degrade when confronted with low-resolution images. We find it is because: 1) low-resolution images lack detail information; 2) current methods complete weak global modeling, which make it difficult to extract discriminative features. To alleviate the above issues, we proposed a novel global multiple extraction network (GME-Net) for low-resolution facial expression recognition, which incorporates 1) a hybrid attention-based local feature extraction module with attention similarity knowledge distillation to learn image details from high-resolution network; 2) a multi-scale global feature extraction module with quasi-symmetric structure to mitigate the influence of local image noise and facilitate capturing global image features. As a result, our GME-Net is capable of extracting expression-related discriminative features. Extensive experiments conducted on several widely-used datasets demonstrate that the proposed GME-Net can better recognize low-resolution facial expression and obtain superior performance than existing solutions.
\end{abstract}

%% file: sec/1_intro.tex
\section{Introduction}
\label{sec:intro}

Facial expression recognition has emerged as a prominent research area in computer vision, attracting extensive attention due to its wide-ranging applications in areas such as human-computer interaction, school education, and monitoring security. In practical scenarios, factors such as camera equipment quality, shooting distance, and image transmission often result in the acquisition of low-resolution face images. These low-resolution images typically lack sufficient facial details, making it challenging to accurately capture and recognize facial expressions.

\begin{figure*}[t]
	\centering
	\includegraphics[width=0.35\textwidth]{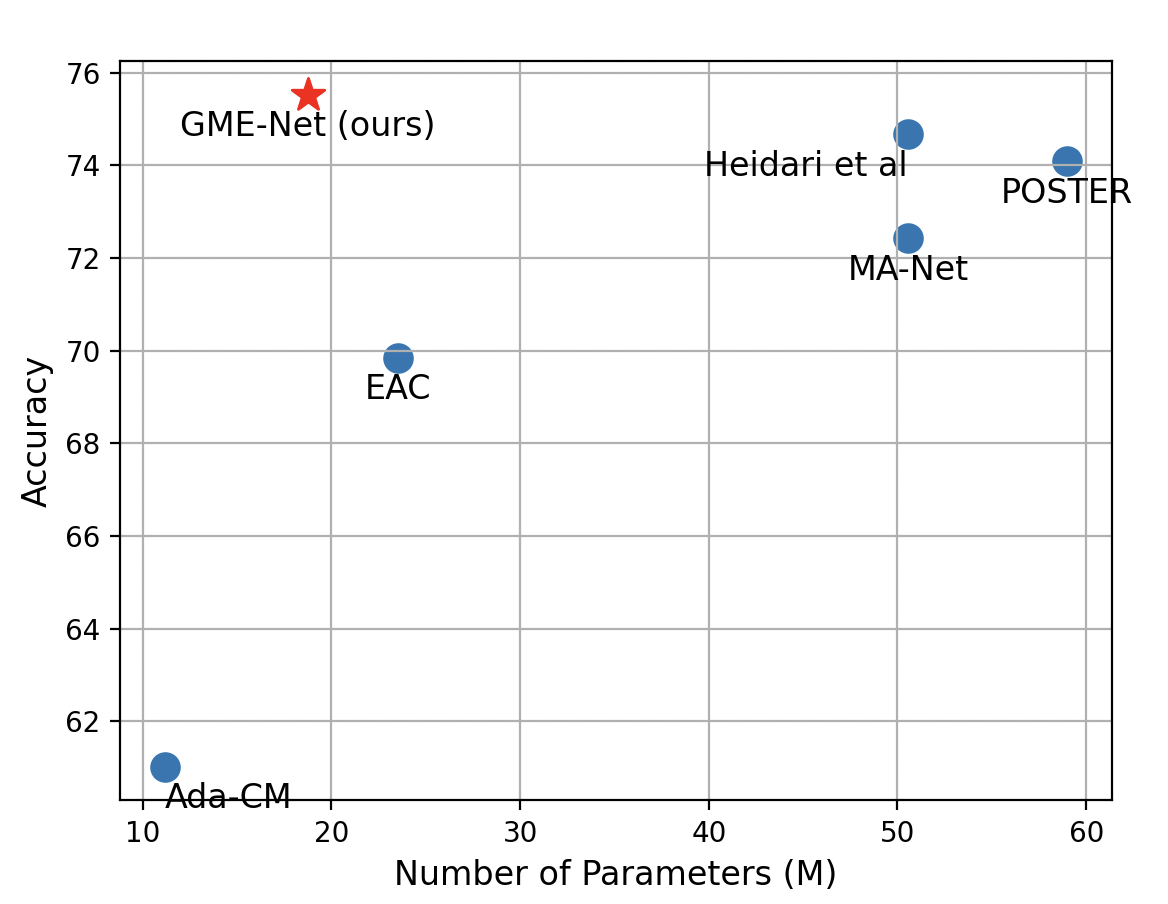}
 \hspace{20mm}
 % \vspace{3mm}
	\includegraphics[width=0.35\textwidth]{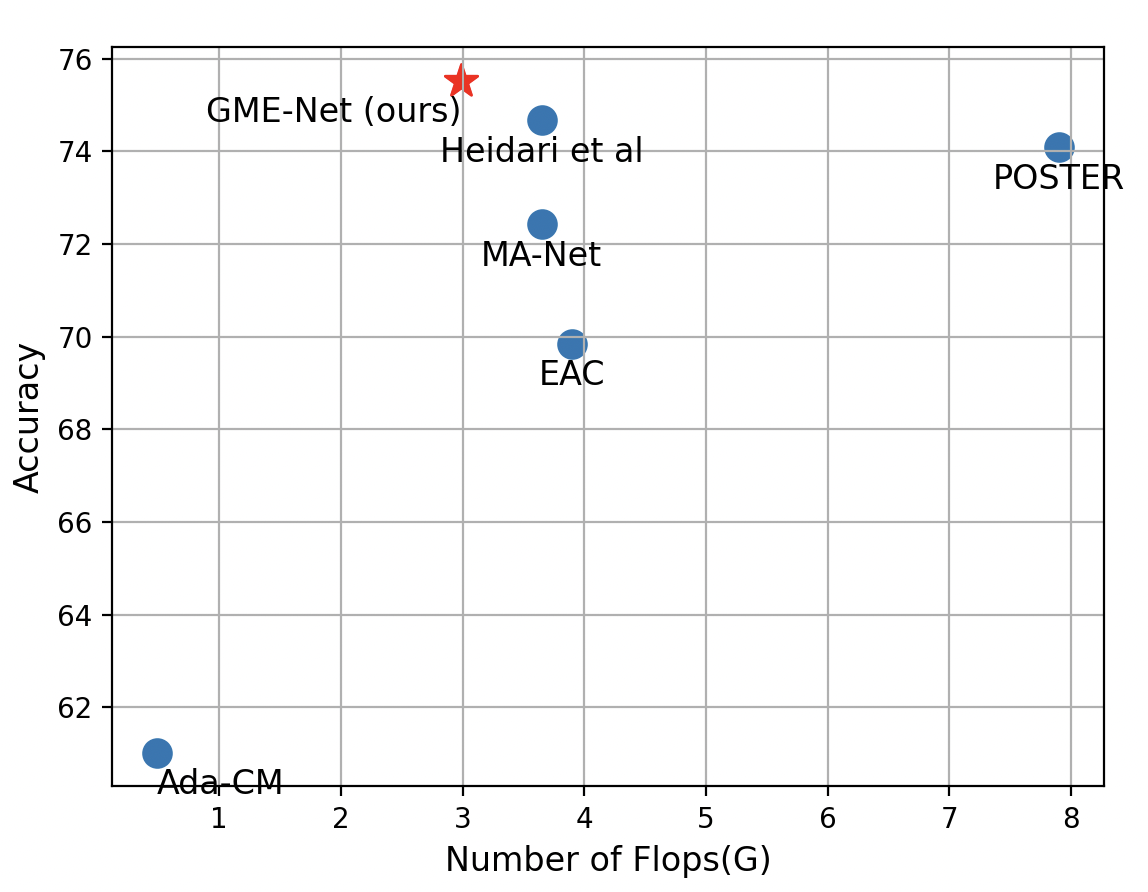}
	\caption{Performance comparison between our method and other FER methods in terms of Accuracy,  computational complexity (GFLOPs),and model parameters on the low-resolution RAF-DB Dataset. In both graphs, our method outperforms the others by achieving the highest accuracy while maintaining a reasonable balance of model complexity and computational cost. This showcases the efficiency and effectiveness of our GME-Net for low resolution facial expression recognition.}
 \vspace{-2mm}
 \label{1}
\end{figure*}

Over the years, researchers have proposed numerous techniques for facial expression recognition. Initial methods employed hand-crafted features and shallow learning techniques such as Local Binary Patterns (LBP) \cite{SHAN2009803}, Histogram of Oriented Gradients (HOG)\cite{1467360}, Gabor\cite{670949}, Non-negative Matrix Factorization (NMF)\cite{5447650}, and Sparse Learning\cite{6247974}. Advancements in deep learning led to the development of facial expression recognition technologies based on Convolutional Neural Network (CNN)\cite{10.1145/2818346.2830587}, Recurrent Neural Network (RNN)\cite{8639182}, and Vision Transformer\cite{Ma_2023}. These deep learning methods demonstrate impressive results on high-resolution images by utilizing large amounts of high-quality data and complex network structures to accurately capture and analyze intricate facial features, enabling precise expression classification. However, their performance tends to degrade when confronted with low-resolution images due to the reduced amount of facial information and semantic details. The experimental results presented in Figure \ref{1} highlight the limitations of existing high-resolution expression recognition methods when applied to a dataset with a resolution of 14x14. Specifically, these methods suffer from low accuracy and large amount of calculation.

Recognizing facial expressions in low-resolution images remains a challenging task, necessitating specific solutions tailored to the low-resolution scenario. Limited studies have explored this direction using various approaches. Ma \textit{et al}. \cite{MA2022108136} utilized a multi-level knowledge distillation technique for low-resolution expression recognition, while Nan \textit{et al}. \cite{NAN2022107678} employed a feature super-resolution method. Additionally, Yan \textit{et al}. \cite{YAN2020107370} proposed a filter learning-based approach. Nevertheless, the achieved results have not yet reached the desired level of accuracy and, in some cases, even fall short of the performance of recognition methods designed for high-resolution images \cite{MA2022108136}.

In related fields such as face recognition, extensive research has been conducted on low-resolution face-related challenges, offering valuable insights into low-resolution facial expression recognition. Some methods aim to obtain high-resolution images by reconstructing details before conducting face recognition, using techniques such as image super-resolution \cite{8100053,7036137,Yang2015RecognitionAA}. These approaches establish a mapping between high-resolution and low-resolution images by designing parameter functions like nonlinear Lagrangian \cite{7299121} and sparse representation \cite{Yang2015RecognitionAA}. While these methods can enhance recognition accuracy, they introduce high computational costs, potentially reducing recognition speed. Alternatively, knowledge distillation techniques \cite{9098036,10.1007/978-3-031-19775-8_37} leverage teacher networks to transfer facial details to student networks, enhancing low-resolution networks' recognition accuracy. 

Inspired by these insights, we propose a novel Global Multiple Extraction Network (GME-Net) for low-resolution facial expression recognition, incorporating a hybrid attention-based local feature extraction module with attention similarity knowledge distillation. This module, comprised of multiple Mixed-Attention Blocks (MAB) with the Depthwise Block Attention Mechanism (DBAM), effectively captures deep facial features and generates expression-related attention maps. By transferring this knowledge from a high-resolution network to a low-resolution network, we provide valuable prior information for accurate expression judgment, guiding the network to focus on the most relevant features.

Additionally, existing facial expression recognition methods often overlook the importance of capturing global features, thereby limiting their performance \cite{MA2022108136,9423267}. To address this limitation, we introduce a multi-scale global feature extraction module consisting of Mixed-Channel Feature Extraction Blocks (MCB). Drawing inspiration from methods \cite{9474949}, MCB is specifically designed to capture expression information from multiple scales while preserving original features to a greater extent. This design approach prevents the network from focusing excessively on local details, thereby mitigating issues of increased intra-class distance and reduced inter-class distance caused by factors such as head posture and face occlusion. Combined with the hybrid attention-based local feature extraction module, our GME-Net integrates features of different scales to obtain global features while maximizing the ability to obtain detailed information using the knowledge distillation framework. In summary, our work makes the following contributions to the field:
\begin{itemize}
% \vspace{-2mm}
\item Our proposed Global Multiple Extraction Net (GME-Net) is evaluated against other methods using the same experimental conditions, demonstrating remarkable performance for low-resolution facial expression recognition.

\item To address the issue of missing facial details in low-resolution images, we propose a hybrid attention-based local feature extraction module, which improves attention consistency between high- and low-resolution networks, enhancing low-resolution expression recognition performance.

\item To mitigate the influence of local noise and capture overall patterns and regularities of facial expressions, we incorporate a multi-scale global feature extraction module into our framework, effectively capturing global features and comprehensively extracting pixel correlations within the image.

\item We generate datasets for low-resolution facial expression recognition due to the lack of publicly available datasets for this specific task, downscaling high-resolution facial expression images from existing datasets as Figure \ref{2} showsz. These datasets provide an opportunity to assess and enhance the performance of low-resolution facial expression recognition methods.
\end{itemize}

%% file: sec/2_formatting.tex
\section{Related Work}
\label{sec:formatting}

In this section, we start by reviewing the current progress in facial expression recognition technology and knowledge distillation techniques. We then discuss relevant literature that applies knowledge distillation methods in the recognition field, including low-resolution expression recognition and related domains like low-resolution face recognition.

%-------------------------------------------------------------------------
\begin{figure}[t]
\scriptsize
\centering
\resizebox{\linewidth}{!}{%
\hspace{-3mm}%
% \begin{adjustbox}{}
\begin{tabular}{cccc}
\rotatebox{90}{\footnotesize{\hspace{6mm}HR image}} &
\includegraphics[height=0.1\textwidth]{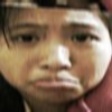}\hspace{-2mm}&
\includegraphics[height=0.1\textwidth]{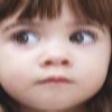}\hspace{-2mm}&
\includegraphics[height=0.1\textwidth]{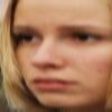}\hspace{-2mm}
\\
\rotatebox{90}{\footnotesize{\hspace{6mm}LR image}}&
\includegraphics[height=0.1\textwidth]{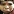}\hspace{-2mm}&
\includegraphics[height=0.1\textwidth]{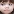}\hspace{-2mm}&
\includegraphics[height=0.1\textwidth]{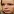}\hspace{-2mm}
\\
\end{tabular}
% \end{adjustbox}
}
\caption{The example of our data set is shown in the figure above. Specifically, it is based on the public data set RAF-DB through the bicubic interpolation method, and the rest of the data set production methods are the same as above.}
% \hspace{-20mm}
% \vspace{-10mm}
\vspace{-4mm}
\label{2}
\end{figure}

%-------------------------------------------------------------------------
\subsection{Facial expression recognition}

In the early stages of expression recognition, traditional machine learning methods were predominantly used, which involved manual feature extraction through the design of feature extraction algorithms. Commonly employed feature extraction methods include HOG \cite{1467360}, LBP \cite{SHAN2009803}, Gabor \cite{670949}, and SIFT \cite{10.1023/B:VISI.0000029664.99615.94}.

With the advancement of expression recognition competitions in recent years, researchers have increasingly focused on facial expressions in wild scenarios, leading to the development of several large-scale facial expression recognition datasets, such as AffectNet \cite{10.1109/TAFFC.2017.2740923}, RAF-DB \cite{8099760}, and FERPlus \cite{10.1145/2993148.2993165}. Deep learning techniques have played a pivotal role in achieving significant advancements in the field of facial expression recognition, with models like AlexNet\cite{10.1145/3065386}, VGGNet\cite{DBLP:journals/corr/SimonyanZ14a}, Inception Net \cite{7298594}, and ResNet \cite{7780459} being widely employed. \cite{DBLP:journals/corr/abs-2109-07270} proposed approaches have demonstrated improved recognition accuracy by maximizing class separability and constructing attention maps for multiple facial regions. To address challenges related to occlusion and pose variance, a regional attention network has been proposed\cite{8974606}. Additionally, attention mechanisms have been integrated into CNN networks, such as pACNN and gACNN\cite{8576656}, to handle occluded facial parts and emphasize features crucial for expression recognition. However, despite these advancements, datasets collected in real-world settings still face challenges such as category imbalance and inaccurate labeling. To mitigate these issues, researchers have employed techniques like the Meta-Face2Exp framework to tackle category imbalance using large-scale face recognition datasets\cite{9879702}. Another approach\cite{9157210} assigns weights to each image and suppresses noisy samples by relabeling labels. Furthermore, attention-consistent erasure methods\cite{10.1007/978-3-031-19809-0_24} have been proposed to prevent the model from overfitting noisy samples.

%-------------------------------------------------------------------------
\subsection{Knowledge Distillation}

Knowledge distillation, originally introduced by Hinton \textit{et al}. \cite{DBLP:journals/corr/HintonVD15}, is a technique that transfers knowledge from complex, high-performance models to smaller models that are more suitable for deployment. It is commonly known as the "teacher-student" training paradigm, where the larger, more complex model acts as the teacher and the smaller model as the student. Knowledge transfer can occur through different approaches, including result-based, feature-based, and relation-based methods. These techniques enable effective knowledge transfer and enhance the performance of the student model.

In terms of result-based knowledge distillation, Zhang \textit{et al}. \cite{8578552} introduced a training approach where multiple student networks are simultaneously trained. The outputs of these networks are used for mutual supervision and guidance, enhancing the learning process. Furlanello \textit{et al}. \cite{DBLP:conf/icml/FurlanelloLTIA18} proposed a regeneration network called Born Again Neural Networks (BAN). It involves training a teacher network and using the same network structure for the student model. The student network progressively replaces the teacher network, iterating until no further improvement is observed, and then integrating all the student networks. Passalis \textit{et al}. \cite{9104915} introduced a probability distribution learning method where the knowledge in the teacher model is represented using probability distributions. The approach involves minimizing the divergence between the probability distributions of the teacher model and the student model, facilitating effective knowledge transfer. For knowledge distillation based on intermediate features, Romero \textit{et al}. \cite{DBLP:journals/corr/RomeroBKCGB14} first proposed a distillation method (FitNets) for learning the eigenvalues of the intermediate layer. The student model uses the advantage of depth to make the performance exceed the teacher network with fewer parameters than the teacher network. Another approach proposed by \cite{8953752} enables the student network to learn not only the output results of the teacher network but also the knowledge of the teacher network's middle layer using a distillation loss function. In \cite{DBLP:conf/iclr/ZagoruykoK17}, a distillation method based on attention transfer (AT) is proposed. This approach utilizes the attention feature maps from the middle layer as the guiding features, enabling the student network to mimic the attention map of the teacher network and enhance its performance. Regarding the knowledge distillation of relational information, Park \textit{et al}. \cite{DBLP:conf/cvpr/ParkKLC19} introduced a distillation loss based on distance and angle, leveraging the relationship between data instances to facilitate the transfer of structural knowledge. The CCKD method, proposed by Peng \textit{et al}.  \cite{DBLP:conf/iccv/PengJLZWLZ019}, not only emphasizes the consistency between instances of teacher and student networks but also highlights the consistency among multiple instances.
%-------------------------------------------------------------------------
\subsection{The Application of KD in LR Image Recognition}

Currently, there is limited research applying knowledge distillation to low-resolution facial expression recognition. Ma \textit{et al}. \cite{10.1016/j.knosys.2022.108136} employed a feature-based knowledge transfer approach. This method utilized the multi-layer features of the teacher network to guide the single-layer output of the student network, assigning different weights to various layer features of the teacher network. In other low-resolution visual recognition tasks like low-resolution face recognition and object recognition, knowledge distillation methods have been employed to address these challenges. Ge \textit{et al}. \cite{DBLP:conf/aaai/GeZLHZJW20} introduced a hybrid sequential relational knowledge distillation method to extract multi-order relational knowledge for image recognition. Zhu \textit{et al}. \cite{8682926} improved the recognition accuracy of the low-resolution network by minimizing the Euclidean distance and cross-entropy loss based on features from both the high-resolution and low-resolution models. Ge  \textit{et al}. \cite{10.1109/TIP.2018.2883743} proposed a selective knowledge distillation approach, where the student network selectively extracts features from the teacher network. \cite{10.1007/978-3-031-19775-8_37} designed knowledge as an attention map to enhance the student network's performance by increasing attention similarity between the teacher and student networks. Soon they presented a feature similarity-based knowledge distillation method\cite{DBLP:journals/corr/abs-2303-04681}.

%% file: sec/3_finalcopy.tex
\section{Method}

In this section, we present GME-Net, which consists of two key modules: the hybrid attention-based local feature extraction module and the multi-scale global feature extraction module. We first provide an overview of the overall architecture of GME-Net and then delve into the details of these two modules.

\subsection{Overall Architecture}

\begin{figure}[t]
	\centering
	\includegraphics[width=0.45\textwidth]{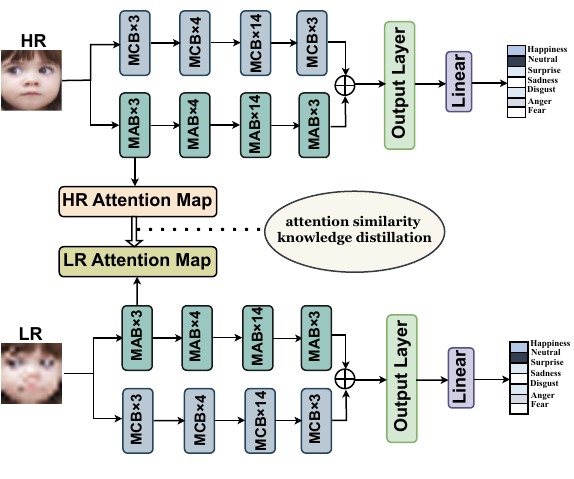}
	\caption{The overall framework of GME-Net, where MAB stands for mixed-attention block, and MCB stands for mixed channel feature extraction block. At the same time, in each MAB we extract an attention map to calculate the distillation loss.}
	\label{3}
 \vspace{-4 mm}
 % \label{Regformer}
\end{figure}

As depicted in Figure \ref{3}, our knowledge distillation framework comprises two components: the high-resolution facial expression recognition network (HR-Net) and the low-resolution facial expression recognition network (LR-Net). In this framework, the teacher network is trained on high-resolution face images, and the student network is trained on low-resolution face images. Additionally, when inputting low-resolution images into the student network, we adjust the size of the image to fit the network input using an interpolation function, and improve the photo quality through Gaussian blur.

In the context of disparate image resolutions, it is difficult for the feature representations of the teacher network and the student network to be completely consistent. To address this issue, taking inspiration from \cite{10.1007/978-3-031-19775-8_37}, we leverage attention maps generated by the teacher network on high-resolution images to guide the student network in focusing on expression-related key parts. In conventional knowledge distillation, the teacher network and the student network are typically distinct, with the former being a larger and more complex model, while the latter is designed to be lightweight. However, in our proposed knowledge distillation framework, the teacher network and the student network share the same architecture. We hope that by sharing the same network structure, the feature representation capabilities between high-resolution and low-resolution networks will be more consistent.

The proposed expression recognition network comprises two branches that work together to enhance the performance of expression recognition. The first branch is a local feature extraction module based on mixed attention, which consists of multiple mixed-attention blocks (MABs). These blocks utilize an attention mechanism to extract crucial local features. The second branch is the multi-scale global feature extraction module, which incorporates multiple Mixed-Channel Feature Extraction Blocks (MCBs). These blocks operate at various scales to capture features at different levels. By combining both local and global features, we can effectively extract global contextual information and alleviate the issue of excessive emphasis on local features, which could overlook overall relevance. The outputs of these two modules are combined through point-wise addition to obtain a fused feature map. Finally, this fused feature map is sent to the output layer for classification to obtain the expression recognition result.

\subsection{Hybrid Attention-based Local Feature Extraction Module}

\begin{figure}[t]
	\centering
 \hspace{-5mm}
	\includegraphics[width=0.45\textwidth]{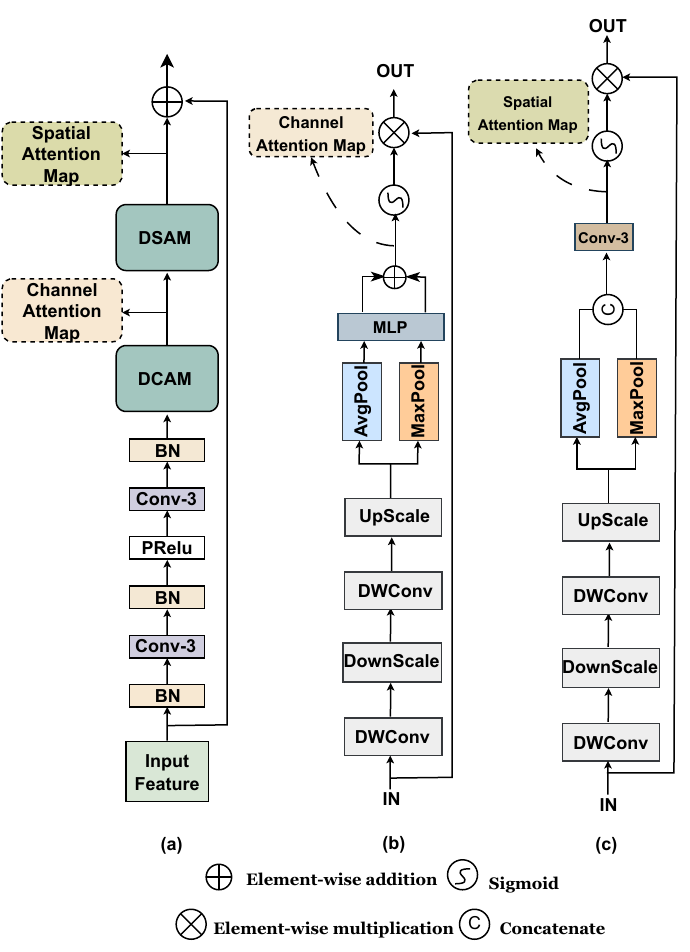}
	\caption{Sub-figures (a) depict the Mixed-Attention Block (MAB), while Sub-figures (b) and Sub-figures (c) illustrate the Depthwise Block Attention Mechanism (DBAM), with Sub-figures (b) representing the Depthwise-Channel Attention Module (DCAM), and Sub-figures (c) denoting the Depthwise-Spatial Attention Module (DSAM).}
	\label{4}
 \vspace{-5 mm}
 % \label{Regformer}
\end{figure}

The ResNet-50 network is employed as the backbone in this module due to its remarkable performance in image recognition tasks. To maintain a simple network structure, we utilize the basicblock residual block in Mixed-Attention Block, which comprises two $3 \times 3$ convolution kernels. To further enhance the network's expressive power and prioritize important features, we introduce our designed Depthwise Block Attention Mechanism (DBAM) based on CBAM \cite{DBLP:conf/eccv/WooPLK18} before the residual connection. The DBAM module combines channel attention and spatial attention mechanisms to make the network more focused on key features, improving the network's perception of important features. Figure \ref{4} shows the structure of the proposed block, the overall process can be expressed as:
\begin{equation}
O=DSAM\left ( DCAM\left ( Conv_{3\times 3} \left (Conv_{3\times 3}\left ( F \right )   \right )  \right )  \right )\oplus F.
\end{equation}
where $F$ denotes the input feature map to the module, $ Conv_{3\times 3}$ refers to the utilization of a $3x3$ convolution operation, $DCAM$ represents the Depthwise-Channel Attention Module that we have designed, $DSAM$ represents the Depthwise-Spatial Attention Module, and $O$ represents the resulting output feature map.

%--------------------------

%------------------------------

\textbf{Depthwise-Channel Attention Module. }  The pooling operation can result in the loss of detailed information, thereby diminishing the model's predictive capability. To maximize the extraction of feature details, we incorporate two depthwise separable convolutions before conducting average pooling and max pooling operations. This process involves a series of $1 \times 1$ depth-separable convolutions, followed by downsampling, a set of $1 \times 1$ depth-separable convolutions, and upsampling. At this point, assuming we have obtained the feature $f$ with dimensions $H \times W \times C$, where $H$, $W$ and $C$ represent the height, width and channel number of the feature. we apply both average pooling and maximum pooling to $f$, followed by a shared fully connected layer. The resulting values are summed element-wise to generate a channel attention map of size $C \times 1 \times 1$ , denoted as $ M_{c} $ . Subsequently, we activate this map using a sigmoid function, producing attention weights ranging from 0 to 1. Finally, we multiply these weights with the input feature map $ F_{c} $, yielding an attention-enhanced feature map that facilitates the network in filtering out valuable channel information. DCAM can be described by the following formula:
\begin{equation}
% \begin{aligned}
\left\{
\begin{array}{lll}
&O_{c} =\sigma \left ( M_{c}  \right ) \otimes F_{c},\\
&F_{m}  = DWConv\left ( DWConv\left ( F_{c}  \right )  \right ), \\
&M_{c} = MLP\left (AvgP\left (  F_{m} \right )   \right ) +  MLP\left (MaxP\left (  F_{m} \right )   \right ),
\end{array}
\right.
% \end{aligned}
\end{equation}
where $F_{c}$ represents the feature map input to the DCAM module; $\sigma$ denotes the sigmoid activation function; $M_{c}$ refers to the channel attention map; and $O_{c}$ represents the resulting output feature map; 
$DWConv$ represents the depth separable convolution operation; $AvgP$ and $MaxP$ denote the average pooling and maximum pooling operations, respectively; $MLP$ refers to the multi-layer perceptron.

\textbf{Depthwise-Spatial Attention Module. }The processing of the first few steps is similar to that of DCAM. It involves employing two depthwise separable convolutions for feature extraction, followed by average pooling and maximum pooling to yield two feature maps of size $H \times W \times 1$. These two feature maps are concatenated along the channel dimension and subjected to a $3 \times 3$ convolutional layer, resulting in a spatial attention map Ms. Subsequently, the Sigmoid activation function is applied, and the obtained weight is multiplied with the input feature map F to enhance attention towards the target region of interest while attenuating attention towards irrelevant areas. DSAM can be described by the following formula:
\begin{equation}
% \begin{aligned}
\left\{
\begin{array}{lll}
&O_{s} =\sigma \left ( M_{s}  \right ) \otimes F_{s},\\
&F_{m}  = DWConv\left ( DWConv\left ( F_{s}  \right )  \right ), \\
&M_{s} =Conv_{3\times 3} \left ( Concat\left ( AvgP\left ( F_{m}  \right ),MaxP\left ( F_{m}  \right )   \right )  \right ) ,
\end{array}
\right.
% \end{aligned}
\end{equation}
where $F_{s}$ represents the feature map input to the DSAM module; $\sigma$ denotes the sigmoid activation function; $M_{s}$ refers to the Spatial attention map; and $O_{s}$ represents the resulting output feature map;$DWConv$ represents the depth separable convolution operation; the $Concat$ means using concatenation; $AvgP$ and $MaxP$ denote the average pooling and maximum pooling operations, respectively.

The  Deep Block Attention Module combines DCAM and DSAM to generate attention maps at both the channel and spatial levels. The incorporation of depthwise separable convolutions aids in extracting detailed features, thereby enhancing the model's predictive capability, without significantly increasing its complexity or computational burden. This design enable the model to effectively process feature information and improve recognition performance in low-resolution facial expression recognition tasks.

\subsection{Multi-scale Global Feature Extraction Module}

\begin{figure}[t]
	\centering
	\includegraphics[width=0.45\textwidth]{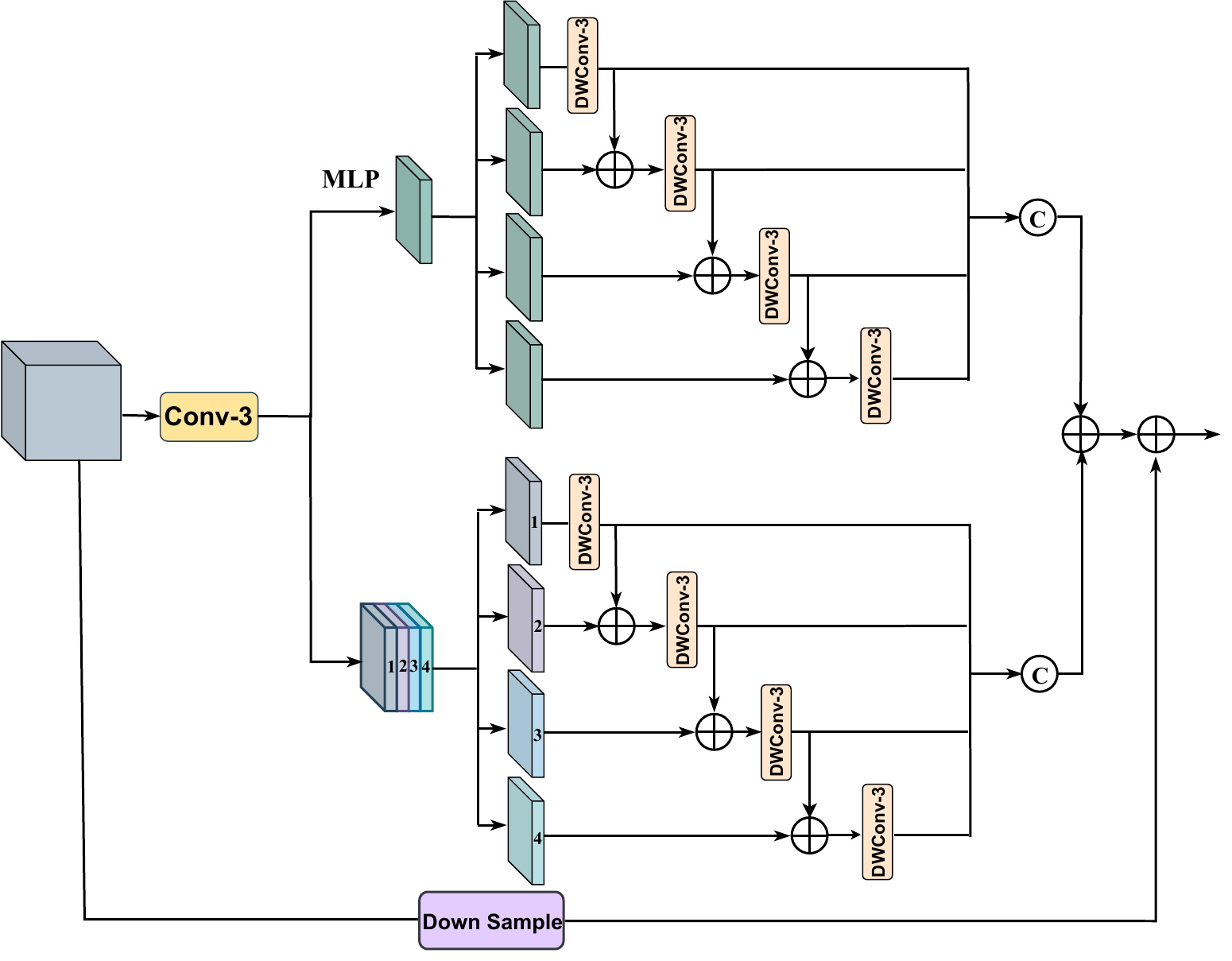}
	\caption{ The structure of Mixed-Channel Feature Extraction Block(MCB).}
	\label{5}
 \vspace{-3mm}
 % \label{Regformer}
\end{figure}

Inspired by Res2Net \cite{10.1109/TPAMI.2019.2938758} and MA-Net \cite{9474949}, we propose a Mixed-Channel Feature Extraction Block in the Multi-scale Global Feature Extraction Module to capture global features. Specifically, we perform a $3\times 3$ convolution on the feature map $F$ to obtain a feature representation of $H \times W \times C$. Then, we input the feature maps into two branches, which adopt a Quasi-symmetric structure as illustrated in Figure \ref{5}. In the first branch, we reduce the channel dimension of the feature map to $H \times W \times C/4$ and replicate it into four copies $X_{1}$, $X_{2}$, $X_{3}$, $X_{4}$. A set of depthwise separable convolutions is applied to extract features from $X_{1}$, resulting in output features $F_{X_{1}}$. We then add $F_{X_{1}}$ and $F_{X_{2}}$, pass them through the next set of depth-separable convolutions, and obtain output features $F_{X_{2}}$. This process is repeated several times until all replicas are processed. Finally, the output features $F_{X_{1}}$, $F_{X_{2}}$, $F_{X_{3}}$, and $F_{X_{4}}$ are concatenated. $O_{1}$ represents the output of the first branch. This design aims to preserve the original features to a great extent while processing the global features, compensating for potential feature loss when the second branch splits the channels. It can be expressed by the following formula: 
\begin{equation}
O_{1} =Concat\left ( F_{X_{1} }  ,F_{X_{2} },F_{X_{3} },F_{X_{4} }\right ) ,
\end{equation}
\begin{equation}
% \begin{aligned}
\left\{
\begin{array}{lll}
&F_{X_{1} } = DWConv_{3\times 3} \left ( X_{1}  \right ) ,\\
&F_{X_{i} } = DWConv_{3\times 3} \left ( F_{X_{i-1}} \right ) \oplus X_{i} (2\le i\le4).
\end{array}
\right.
% \end{aligned}
\end{equation}
In the second branch, we partition the feature map into four segments ($Y_{1}$, $Y_{2}$, $Y_{3}$, $Y_{4}$) based on the channel count, and apply the same processing as in the first branch. This yields output features $F_{Y_{1} }$, $F_{Y_{2} }$, $F_{Y_{3} }$, and $F_{Y_{4} }$, which are then concatenated together. $O_{2}$ represents the output of the second branch.
\begin{equation}
O_{2} =Concat\left ( F_{Y_{1} }  ,F_{Y_{2} },F_{Y_{3} },F_{Y_{4} }\right ) ,
\end{equation}
% and
% \begin{aligned}
\begin{equation}
\left\{
\begin{array}{lll}
&F_{Y_{1} } = DWConv_{3\times 3} \left ( Y_{1}  \right ), \\
&F_{Y_{i} } = DWConv_{3\times 3}  ( F_{Y_{i-1}}    ) \oplus Y_{i} (2\le i\le4).
\end{array}
\right.
% \end{aligned}
\end{equation}
Finally, we combine the concatenated results from the two branches and apply a residual connection with the original feature map. This enables us to effectively extract global and local features from multiple scales in a more efficient manner.The final result can be expressed as follows:
\begin{equation}
O=O_{1} +  O_{2}+F,
\end{equation}
where $O$, $O_{1}$, and $O_{2}$ denote the final output of the module, the output of the first branch, and the output of the second branch, respectively.

\subsection{Loss Function}

In 3.2 Local Feature Extraction Module Based on Mixed Attention, we will get channel attention map and spatial attention map. The cosine distance between the attention maps of the teacher network and the student network is calculated according to the following formula:
\begin{equation}
Similarity_{c}=  \frac{M_{C,T}\cdot M_{C,S} }{\left \|M_{C,T}  \right \|_{2}  \left \|M_{C,S}  \right \|_{2}},
\end{equation}
\begin{equation}
Similarity_{s}=  \frac{M_{S,T}\cdot M_{S,S} }{\left \|M_{S,T}  \right \|_{2}  \left \|M_{S,S}  \right \|_{2}},
\end{equation}
where $Similarity_{c}$ represents the cosine similarity of the channel attention maps between the teacher network and the student network, $Similarity_{s}$ represents the cosine similarity of the spatial attention maps between the teacher network and the student network. $M_{C,T}$ denotes the channel attention map of the teacher network, $M_{C,S}$ represents the channel attention map of the student network, while $M_{S,T}$ denotes the spatial attention map of the teacher network and $M_{S,S}$ represents the spatial attention map of the student network, $ \| \cdot \|_{2}$ denotes L2-norm.

We aim to increase the similarity between the attention maps generated by the teacher network and the student network by reducing the cosine distance between them. This approach helps improve the recognition accuracy of the student network, specifically designed for low-resolution face recognition. According to the cosine similarity of the attention map, the cosine distance can be expressed as 1 minus the cosine similarity, then our knowledge distillation loss can be expressed as:
\begin{equation}
L_{k d} =\frac{\left (  1-Similarity_{c}\right ) + \left (  1-Similarity_{s}\right ) }{2}, 
\end{equation}

We compute the distillation loss by taking the average of the channel cosine distance and the spatial cosine distance. Along with the distillation loss, we incorporate the target task loss, which is essential for our objective. For expression recognition, we utilize the widely employed cross-entropy loss function to quantify the disparity between the model's output and the actual label.Therefore, the total loss can be expressed as the weighted sum of the distillation loss and the cross-entropy loss, and the formula is expressed as:
\begin{equation}
L_{ce} =- \frac{1}{N} \sum_{i= 1}^{N} \sum_{j= 1}^{C} y_{i}^{j}\log_{}{\left ( p_{j}\left ( x_{i },\theta  \right )   \right ) }
\end{equation}
\begin{equation}
L=L_{ce}  + \lambda _{kd}   L_{kd} ,
\end{equation}
where $N$ denotes the total number of samples in the dataset, $C$ is the number of expression categories, $ p_{j}\left ( x_{i },\theta  \right )   $ denotes the predicted probability of sample $x_{i}$ belonging to category $j$, $\theta$ represents the model parameter, and $y_{i}^{j}$ represents the corresponding true label value.

%% file: sec/4.tex
\section{Experiments}
\subsection{Experimental Settings}

\begin{table*}
\hspace{-2mm}
\caption{Comparing with state-of-the-art methods on the low-resolution RAF-DB dataset (14x14 resolution). 'train with lr' indicates that the method is trained on a low-resolution dataset and tested on a low-resolution dataset; 'train with hr' indicates that the method is trained on a high-resolution dataset and tested on a low-resolution dataset;' train with hr+lr' means to train the method with high-resolution and low-resolution data sets, and then test on the low-resolution data set.}
\scalebox{1.02}{
\begin{tabular}{c|c|ccc|c|c}
\hline
Methods       & Years & train with lr & train with hr & train with lr+hr & Number of Parameters(M) & Number of Flops(G) \\
\hline
MA-Net        & 2021  & 70.27         & 60.27         & 72.43            &50.55                      & 3.65               \\
Ada-cm        & 2022  & 61.02         & 55.67         & 59.32            & \textbf{11.18}                 & \textbf{0.49}      \\
EAC           & 2022  & 66.07         & 62.68         & 69.85            & 23.52                    & 3.90               \\
POSTER        & 2022  & 72.07         & 68.12         & \textcolor{cyan}{74.09}            & 58.98                    & 7.90               \\
Heidari \textit{et al}& 2022  & 73.44         & 65.06         & \textcolor{blue}{74.67}            & 50.55                     & 3.65               \\
\hline
GME-Net(ours) & 2023     &  \multicolumn{3}{c|}{\textcolor{red}{75.52}}               & \textbf{18.75}                     & \textbf{2.99} \\    
\hline
\end{tabular}
}
\label{6}
\end{table*}

\textbf{Evaluated datasets.}The knowledge distillation framework we employ requires feeding high-resolution facial expression images and low-resolution facial expression images to the teacher network and student network, respectively. Since there is currently no public low-resolution facial expression recognition dataset, we generate a suitable low-resolution facial expression recognition dataset based on the existing high-resolution facial expression dataset to facilitate our model training process. We selected several widely-used benchmarks, namely RAF-DB \cite{8099760}, ExpW \cite{DBLP:journals/ijcv/ZhangLLT18}, FER2013 \cite{DBLP:conf/iconip/GoodfellowECCMHCTTLZRFLWASMPIPGBXRXZB13}, and FERPlus \cite{10.1145/2993148.2993165} , which consist of real-world facial expression images, to evaluate the performance of our model. To simulate the low-resolution scenario encountered in practical situations, we downscaled the images using the Bicubic interpolation method at various downsampling ratios.

1) RAF-DB \cite{8099760}: RAF-DB is a real-world dataset obtained from the internet, containing nearly 30,000 facial images annotated by 40 annotators. In our experiments, we chose single-label subsets featuring seven basic expressions, which were divided into training and test sets, consisting of 12,271 and 3,068 images, respectively. We downsampled this dataset to a resolution of 14x14.

2) ExpW \cite{DBLP:journals/ijcv/ZhangLLT18}: The ExpW dataset consists of 91,793 facial images sourced from Google Image Search. These images have been manually annotated into seven basic expression categories. To address issues with the original data quality, we conducted preprocessing on the experimental data, including facial landmark detection, face alignment, and removal of non-face images. This resulted in a final collection of 87,305 facial images with a resolution of 112x112. Based on the distribution of expression types, we designated 10\% of the dataset as the test set, while the remaining 90\% serves as the training set. We reduced the resolution of this dataset to 14x14.

3)FER2013 \cite{DBLP:conf/iconip/GoodfellowECCMHCTTLZRFLWASMPIPGBXRXZB13} The FER2013 (Facial Expression Recognition 2013) dataset is a widely used dataset for facial expression recognition. It contains 35,887 grayscale facial images, with each image sized at 48x48 pixels. These images are divided into seven categories, namely: Angry, Disgust, Fear, Happy, Sad, Surprise, and Neutral. The FER2013 dataset was collected through the internet, and each image has been annotated by one or more human annotators. We downsampled this dataset to a resolution of 12x12.

4)FERPlus \cite{10.1145/2993148.2993165} : The FERPlus is an extension of the original FER2013 dataset, where the images have been re-labelled into one of 8 emotion types: neutral, happiness, surprise, sadness, anger, disgust, fear, and contempt.This dataset engages more human annotators to label images and introduces a multi-label classification system, allowing an image to contain multiple expressions. This increased complexity enhances the dataset's ability to tackle real-world facial expression recognition challenges. This dataset was downsampled to a 12x12 resolution.

\vspace{1 mm} 
\textbf{Compared methods.}Considering the limited availability of low-resolution facial expression recognition methods, and the fact that most code and low-resolution datasets used are not open-sourced, it is challenging to make a fair comparison. Therefore, we opt to compare our approach with state-of-the-art high-resolution methods on images to emphasize the advantages our method offers over them. We have selected facial expression recognition techniques from the past two years, including Ada-CM \cite{9880204}, MA-Net \cite{9474949}, Poster \cite{DBLP:journals/corr/abs-2204-04083}, EAC \cite{10.1007/978-3-031-19809-0_24}, and Diversified-fer \cite{heidari2023learning}. We train and test these methods on the dataset we produced, adhering to their original experimental settings.

\vspace{1 mm} 
\textbf{Implementation and training details.} In our GME-Net, we set the initial number of channels to 32, and the weight factor for distillation is set to 5. For training, we utilize the SGD optimizer with a momentum of 0.9 and an initial learning rate of 0.1. The learning rate is multiplied by 0.4 every 20 epochs. We use a training batch size of 64, and the total number of epochs is 100. The training process is conducted on NVIDIA GeForce RTX 3090.

\begin{table}
\hspace{-2.5mm}
\caption{Comparing with state-of-the-art methods on the low-resolution FerPlus dataset (12x12 resolution).}
\scalebox{0.9}{
\begin{tabular}{c|ccc}

\hline
Methods       & train with lr & train with hr & train with lr+hr \\
\hline
MA-Net        & 70.36         & 43.28         & \textcolor{blue}{70.75}   \\
Ada-CM       & 47.08         & 35.78         & 44.11            \\
EAC           & 64.79         & 48.26         & 66.98            \\
POSTER        & 68.47         & 52.21      & {70.01}            \\
Heidari \textit{et al} & 69.45         & 49.18         & \textcolor{red}{71.01}   \\
\hline
GME-Net(ours) & \multicolumn{3}{c}{\textcolor{cyan}{70.57}} \\
\hline
\end{tabular}
}
\label{7}
\end{table}

\subsection{LR-FER Performance Evaluations} 
In this section, due to the limited methods available for low-resolution facial expression recognition and the fact that most of the codes and low-resolution datasets used are not publicly available, we compare our approach with several state-of-the-art methods commonly used for high-resolution datasets. We trained, tested, and compared all methods on a self-constructed low-resolution dataset based on the RAF-DB dataset, FerPlus dataset and ExpW dataset to evaluate their performance. It is important to highlight that our method was not pretrained on large-scale datasets. Consequently, the comparison methods in our study also did not employ pretrained models. In addition, to ensure fair comparative experiments, we took into account the unique nature of the knowledge distillation framework and the utilization of both high-resolution and low-resolution datasets. When assessing the performance of other methods, we divided each experimental group into three versions: training with low-resolution facial images, training with high-resolution facial images, and training with both low-resolution and high-resolution datasets. All three versions underwent testing on the low-resolution dataset, thereby maintaining consistency and fairness in the comparisons.

\begin{table}
\hspace{-2.5mm}
\caption{Comparing with state-of-the-art methods on the low-resolution ExpW dataset (14x14 resolution).}
\scalebox{0.9}{
\begin{tabular}{c|ccc}
\hline
Methods       & train with lr & train with hr & train with lr+hr \\
\hline
MA-Net        & 64.19         & 54.92         & \textcolor{blue}{66.56}   \\
Ada-CM        & 30.76         & 37.30         & 31.72            \\
EAC           & 64.97         & 55.75         & 65.21            \\
POSTER        & 64.44         & 55.80         & 64.85            \\
Heidari \textit{et al} & 65.21         & 56.46         & \textcolor{cyan}{65.70}   \\
\hline
GME-Net(ours) & \multicolumn{3}{c}{\textcolor{red}{67.45}} \\
\hline
\end{tabular}
}
\label{8}
\end{table}

1) Comparison on low-resolution RAF-DB. As indicated in Table \ref{6}, our method achieves an accuracy rate of 75.52\% on the 14x14 low-resolution RAF-DB dataset, which demonstrates its high competitiveness. Our results outperform Ada-cm and EAC methods by a significant margin, with a 3.09\% higher accuracy compared to MA-Net, a 1.43\% higher accuracy compared to POSTER, and a 0.85\% higher accuracy compared to the method proposed by Heidari et al.. At the same time, it can be observed that other methods tend to achieve the highest accuracy when training with both low-resolution and high-resolution datasets together. Conversely, when training solely with high-resolution datasets, the obtained results for testing on low-resolution images are comparatively lower. 

2) Comparison on low-resolution FerPlus. As shown in Table \ref{7}, the method proposed by Heidari et al. achieved the highest accuracy rate of 71.01\%, followed by MA-Net which achieved 70.75\%, and our proposed method was slightly lower than them, reaching 70.57\%. But it is worth noting that, as shown in Table \ref{6}, the parameters and flops of the top two methods are much higher than our method.

3) Comparison on low-resolution ExpW. As shown in Table \ref{8}, our method achieved an accuracy rate of 67.45\% on the EXPW dataset, which is 0.89\% higher than the MA-Net method, 1.75\% higher than the method proposed by Heidari et al. 2.6\%, 2.24\% higher than EAC.

\subsection{ Ablation Studies} 
To assess the effectiveness of each module in our GME-Net, we conducted a comprehensive ablation analysis. For this purpose, we selected multiple datasets as evaluation benchmarks, allowing us to thoroughly evaluate the performance of each module. 
\begin{table}
\caption{Ablation study on low-resolution RAF-DB Dataset(14x14 resolution) and low-resolution FER2013 DataSet(12x12 resolution). It reflects the role of each component in our GME-Net.}
\begin{tabular}{c|cc}
\hline
Methods                      & RAF-DB           & FER2013                    \\
\hline
Baseline(Resnet-50)          & 71.0654          & 50.1254                   \\
Baseline+CBAM                & 73.7288          & 52.5216                 \\
Baseline+DBAM                & 74.2940          & 54.7506                  \\
Baseline+Global Module       & 71.8383          & 50.3283                    \\
Baseline+DBAM+GM(without kd) & 71.5361          & 50.9613                  \\
Baseline+DBAM+GM(GME-Net)& \textbf{75.5215} & \textbf{56.6174} \\
\hline
\end{tabular}
\label{9}
\end{table}

As shown in the Table \ref{9} and Table \ref{10} ,  we present the results of our ablation analysis. The baseline model is Resnet-50. "Baseline+CBAM" refers to the baseline model with the addition of the Convolutional Block Attention Module (CBAM)\cite{DBLP:conf/eccv/WooPLK18}. "Baseline+DBAM" indicates the baseline model enhanced with our Depthwise Block Attention Mechanism (DBAM) . "Baseline+Global Module" includes a Multi-scale Global Feature Extraction Module added to the baseline model. "Baseline+DBAM+GM (without kd)" incorporates both the DBAM and Global Module without utilizing the knowledge distillation framework, which means it does not use the guidance of the teacher network's attention map. Lastly, "Baseline+DBAM+GM" represents the final version of our method, which includes both modules and utilizes attentional similarity knowledge distillation for high-resolution network knowledge transfer.

1)\textbf{Branch 1 (Hybrid Attention-based Local Feature Extraction Module)}. Branch 1 is a crucial component of GME-Net that enhances the network's capability to extract local features. In order to assess the effectiveness of branch 1, we conducted experiments using ResNet-50 as the baseline model. Initially, we added the CBAM module to the baseline model and evaluated its performance. Subsequently, we replaced the CBAM module with our DBAM module to validate the efficacy of our proposed enhancements.

\begin{table}
\caption{Ablation study on low-resolution ExpW Dataset(14x14 resolution) and low-resolution FERPlus   DataSet(12x12 resolution). It reflects the role of each component in our GME-Net.}
\begin{tabular}{c|cc}
\hline
Methods                              & ExpW             & FERPlus          \\
\hline
Baseline(Resnet-50)         & 64.6136          & 66.4195          \\
Baseline+CBAM               & 65.6898          & 68.0214          \\
Baseline+DBAM               & 66.5942          & 69.8902          \\
Baseline+Global Module       & 64.9685          & 66.5295          \\
Baseline+DBAM+GM(without kd) & 65.1675          & 66.5135          \\
Baseline+DBAM+GM(GME-Net)    & \textbf{67.4528} & \textbf{70.5725}\\
\hline
\end{tabular}
\label{10}
\end{table}

As indicated in Table \ref{9} and Table \ref{10}, our proposed method demonstrates notable improvements in the experiments conducted on the low-resolution-RAF-DB dataset. For images with a resolution of 14×14, our method achieves an accuracy rate that is 4.46\% higher than the baseline model, surpassing the performance achieved by adding the CBAM module, which shows an improvement of 1.79\%. Similarly, on the low-resolution-FER2013 dataset with images of 12×12 resolution, our method achieves an accuracy rate that is 6.49\% higher than the baseline model, surpassing the network with the CBAM module added by 4.10\%. On the low-resolution-ExpW dataset with images of 14×14 resolution, our method achieves an accuracy rate that is 2.84\% higher than the baseline model, and 1.76\% higher than the network with the CBAM module added. Finally, on the low-resolution-FERPlus dataset with images of 12×12 resolution, our method achieves an accuracy rate that is 4.15\% higher than the baseline model and 2.55\% higher than the network with the CBAM module added. These experimental results further validate the effectiveness of branch 1, highlighting the advantages of our improved module in enhancing the network's attention and extraction of local features. This provides strong support for the performance enhancement of our GME-Net in low-resolution facial expression recognition tasks.

2)\textbf{ Branch 2 (Multi-scale Global Feature Extraction Module.)} Then we evaluate the performance of branch 2. This branch is a key module we propose, which is used to introduce a channel hybrid extraction mechanism to extract global and local features from the channel level. In the ablation experiment, we mainly conducted two tests: adding branch 2 to the baseline model and removing the branch 2 module from the complete network.

First, we added branch 2 to the baseline model. By using the same dataset and experimental settings, we compared the performance of the baseline model with the model after adding branch 2. On the low-resolution-RAF-DB dataset, the accuracy rate increased by 0.77\% compared to the baseline. On the low-resolution-FER2013 dataset, the accuracy rate improved by 0.20\% compared to the baseline. On the low-resolution-ExpW dataset, the accuracy rate increased by 0.35\% compared to the baseline. On the low-resolution-FERPlus dataset, the accuracy rate improved by 0.11\% compared to the baseline.

Next, we removed branch 2 from GME-Net,  eliminating the channel mixture extraction mechanism from the complete network. By comparing the performance of the full network with and without branch 2 on the experimental dataset, we can observe the impact of branch 2 on the overall network performance. Based on the results obtained, we can see that on the low-resolution-RAF-DB dataset, the accuracy of GME-Net increased by 1.23\% due to the inclusion of branch 2. Similarly, on the low-resolution-FER2013 dataset, the accuracy rate increased by 1.87\%. On the low-resolution-ExpW dataset, the accuracy rate improved by 0.86\%, and on the low-resolution-FERPlus dataset, the accuracy rate increased by 0.68\%. These data clearly demonstrate the importance of branch 2 in GME-Net and highlight its positive contribution to our network's performance.

3)\textbf{Knowledge distillation method.} Furthermore, we conducted a study on the effectiveness of the knowledge distillation method, which serves as a key approach for knowledge transfer by leveraging the guidance of a teacher network during the training of the student network. We compared with performance of training only the student network  without the knowledge distillation method.

On the low-resolution-RAF-DB dataset, the accuracy rate increased by 3.99\% when employing the knowledge distillation method. Similarly, on the low-resolution-FER2013 dataset, the accuracy rate saw an improvement of 5.66\%. On the low-resolution-ExpW dataset, the accuracy rate increased by 2.29\%, and on the low-resolution-FERPlus dataset, there was a 4.06\% increase in accuracy rate. These experimental findings clearly demonstrate that the knowledge distillation method can significantly enhance the performance of low-resolution networks in expression recognition tasks.

Through the comprehensive analysis of the ablation experiments, we have successfully validated the effectiveness of each module within the GME-Net architecture. The experimental results have provided substantial evidence for the efficacy of branch 1 in enhancing attention, the role of branch 2 in multi-scale global feature extraction, and the advantageous knowledge transfer achieved through the distillation method.

%% file: sec/5.tex
\section{Conclusion}
In conclusion, this research addresses the challenges of low-resolution facial expression recognition by proposing the Global Multiple Extraction Network (GME-Net). The limitations of existing methods, including the lack of detail information in low-resolution images and weak global modeling, are effectively addressed by our approach.  The key contributions of our work include the incorporation of a hybrid attention-based local feature extraction module and a multi-scale global feature extraction module. The hybrid attention-based module leverages attention similarity knowledge distillation to learn image details from a high-resolution network, while the multi-scale global feature extraction module mitigates the impact of local image noise and enhances the capture of global image features. Through extensive experiments on widely-used datasets, our GME-Net demonstrates superior performance in low-resolution facial expression recognition compared to existing solutions. The ability of our network to extract expression-related discriminative features contributes to its effectiveness in addressing the challenges posed by low-resolution images. The proposed GME-Net offers a promising approach for improving the recognition of facial expressions in low-resolution images, thereby advancing the field of computer vision and contributing to the development of more robust facial expression recognition algorithms.In future work, we plan to optimize the model further and address challenges associated with the application of low-resolution facial expression recognition technology in real-world scenarios, such as lighting changes and variations in facial poses.